\documentclass[runningheads]{llncs}
   
\usepackage{graphicx}
\usepackage{textcomp}
\usepackage{array,ragged2e}

\def\BibTeX{{\rm B\kern-.05em{\sc i\kern-.025em b}\kern-.08em
    T\kern-.1667em\lower.7ex\hbox{E}\kern-.125emX}}

\usepackage{colortbl}
\usepackage{makecell}
\usepackage{pgfplots}
\pgfplotsset{compat=1.17}
\usepackage{amsmath}
\usepackage{multicol}
\usepackage{multirow}
\usepackage{tabto}
\usepackage{wrapfig}
\usepackage{amssymb}
\usepackage{amsmath}
\usepackage{float}
\usepackage{graphicx}
\usepackage[misc,geometry]{ifsym}
\usepackage{tabularx}
\usepackage{xtab}
\usepackage{longtable}
\usepackage{graphicx}
\usepackage{caption}
\usepackage{algorithm}
\usepackage{algpseudocode}

\usepackage{textcomp}
\usepackage{array,ragged2e}
\usepackage{pgfplots}
\usepackage{url}
\usepackage{lineno,hyperref}
\hypersetup{
  linkcolor  = red!60!black,
  citecolor  = blue!90!white,
  urlcolor   = violet!60!black,
  colorlinks = true,
}

\definecolor{LightCyan}{rgb}{0.88,1,1}
\definecolor{LightYellow}{rgb}{1,1,0.6}
\usepackage{orcidlink}

\usepackage{hyperref}
\usepackage{svg}
\usepackage{xcolor}
\newcommand{\orcid}[1]{\href{https://orcid.org/#1}{\includesvg[width=10pt]{orcid.svg}}}

\usepackage{threeparttable}

\usepackage{pgf-pie}  

\begin{document}
\title{Enhancing Bagging Ensemble Regression with Data Integration for Time Series-Based \\Diabetes Prediction}

\titlerunning{Predicting Diabetes on Time Series}

\author{Vuong M. Ngo\orcidlink{0000-0002-8793-0504}\inst{1,2}~\textsuperscript{\Letter}
\and
Tran Quang Vinh\orcidlink{0000-0002-6492-4857}\inst{3}~\textsuperscript{\Letter}
\and \\
Patricia Kearney\orcidlink{0000-0001-9599-3540}\inst{4}
\and
Mark Roantree\orcidlink{0000-0002-1329-2570}\inst{1}
}
\authorrunning{Ngo, V.M. et al.}
\institute{Insight Centre for Data Analytics, School of Computing, Dublin City University, Dublin, Ireland \and
Faculty of Information Technology, Ho Chi Minh City Open University, Ho Chi Minh City, Vietnam \and
Faculty of Electrical and Electronic Engineering, Ho Chi Minh City University of Transport, Ho Chi Minh City, Vietnam \and
School of Public Health, University College Cork, Cork, Ireland\\
\email{vuong.ngo@dcu.ie or vuong.nm@ou.edu.vn, vinh.tran@ut.edu.vn, patricia.kearney@ucc.ie, mark.roantree@dcu.ie}
}


\maketitle

\begin{abstract}
Diabetes is a chronic metabolic disease characterized by elevated blood glucose levels, leading to complications like heart disease, kidney failure, and nerve damage. Accurate state-level predictions are vital for effective healthcare planning and targeted interventions, but in many cases, data for necessary analyses are incomplete. This study begins with a data engineering process to integrate diabetes-related datasets from 2011 to 2021 to create a comprehensive feature set. We then introduce an enhanced bagging ensemble regression model (EBMBag+) for time series forecasting to predict diabetes prevalence across U.S. cities. Several baseline models, including SVMReg, BDTree, LSBoost, NN, LSTM, and ERMBag, were evaluated for comparison with our EBMBag+ algorithm. The experimental results demonstrate that EBMBag+ achieved the best performance, with an MAE of 0.41, RMSE of 0.53, MAPE of 4.01, and an R$^2$ of 0.91. 
\end{abstract}


\section{Introduction}
\label{sec:Introduction}
Diabetes is a chronic metabolic disease characterized by elevated blood glucose levels, which can cause severe damage to the heart, blood vessels, eyes, kidneys, and nerves over time. Globally, approximately 830 million people live with diabetes, primarily in low- and middle-income countries, with 1.5 million deaths attributed to the disease annually \cite{WHO-Diabetes-24}. In the U.S., diabetes presents a major public health challenge, affecting 38.4 million people, or 11.6\% of the population, with substantial state-level variations influenced by demographic and economic factors \cite{CDC-Diabetes-24}. Accurate prediction of diabetes prevalence by state is essential for effective healthcare planning, resource allocation, and targeted interventions, especially considering the high rates among seniors (29.2\% or 16.5 million) and youth (0.35\% or 352 thousands under 20). Diabetes also disproportionately affects racial and ethnic minorities, highlighting the need for culturally tailored approaches.

In the U.S., the disease's economic toll is significant, costing \$412.9 billion in 2022, with individuals with diabetes incurring healthcare costs 2.6 times higher than those without \cite{ADS-24}. Given that diabetes was the eighth leading cause of death in 2021, accurate state-level predictions are crucial to reducing the health and economic burden of diabetes nationwide and improving population health outcomes.  



In recent years, data mining and ML have become essential, reliable tools in the medical field. Data mining is used to preprocess healthcare data and select relevant features, while ML automates the prediction of conditions like diabetes. Many studies require careful integration of data that has been sourced from separate, often heterogenous, databases and repositories \cite{10386435}. Then, by uncovering hidden patterns in the data, complex ML methods may enable accurate and reliable decision-making \cite{Khanam-Foo:2021}, \cite{Ganie:2023}. However, recent studies have rarely incorporated time-series features in their models and have primarily focused on predicting diabetes at the individual level rather than at a broader, city-wide scale. Specifically, to improve performance, ML models should be enhanced to better adapt to healthcare applications and disease detection.

Our contribution can be articulated as follows:
\begin{itemize}
    \item Engineering a novel diabetes dataset containing a comprehensive feature set suitable for machine learning algorithms by integrating various U.S. diabetes-related data sources.
    \item The application of time series techniques to several popular ML models to establish baseline models for diabetes prediction.
    \item The development of an enhanced bagging ensemble regression model (EBMBag+) that incorporates time series techniques for predicting diabetes prevalence.
    \item Delivering a robust evaluation framework incorporating the analysis of EBMBag+ against baseline models to demonstrate its superior performance.
\end{itemize}

The remainder of this paper is organized as follows: Section \ref{sec:RW} reviews the related work, while Section \ref{sec:Dataset} presents four time-series datasets along with our integrated diabetes dataset for supervised learning models. Section \ref{sec:Methods} describes our system architecture, the structuring of time-series training and testing datasets, popular time-series prediction models, and our proposed EBMBag+. In Section \ref{sec:exp_analysis}, we evaluate the models using various metrics and provide a detailed analysis. Finally, Section \ref{sec:Conclusion} concludes the paper and offers insights for future research directions.

\section{Related Work}
\label{sec:RW}

Several studies have applied ML algorithms to predict diabetes using patients' medical records, including  \cite{Tasin:2023}, \cite{Orlando:2023}, \cite{Sofany:2024}, \cite{Lugner:2024}, \cite{Dharmarathne:2024} and  \cite{Modak:2024}. In \cite{Tasin:2023}, the authors used the Pima Indian diabetes dataset and collected additional samples from 203 individuals working at a local textile factory in Bangladesh, focusing on six features: pregnancy, glucose level, blood pressure, skin thickness, BMI, age, and diabetes outcome. To address class imbalance in the dataset and enhance ML performance, they applied Synthetic Minority Over-sampling Technique and Adaptive Synthetic Sampling. 

In \cite{Orlando:2023}, the five ML algorithms were employed to classify binary outcomes, focusing specifically on type 2 diabetes. Type 2 diabetes is characterized by the body’s reduced ability to use insulin effectively, often linked to lifestyle factors. The authors used a Kaggle dataset containing 768 patient records with nine features: number of pregnancies (for female patients), glucose level, diastolic blood pressure, skinfold thickness, insulin level, body mass index, family history of diabetes, age, and diabetes outcome (yes/no).

In \cite{Sofany:2024}, the authors proposed using the XGB method in their mobile app to predict diabetes. Their dataset includes 300 data samples from volunteers, collected from specialty hospitals in Saudi Arabia and Egypt during the 2022-2023 academic year. A key contribution of this study is the development of a mobile app that allows users to input relevant features and instantly receive a diabetes prediction.

In \cite{Lugner:2024}, the SHapley Additive Explanation technique was used to identify the most influential factors for predicting the 10-year risk of developing type 2 diabetes, followed by the use of the XGBoost model for diabetes classification. Using a dataset of 12,148 participants with type 2 diabetes, they concentrated on the top ten features associated with diabetes risk. HbA1c emerged as the strongest predictor, followed by BMI, waist circumference, and blood glucose levels.

In \cite{Dharmarathne:2024}, the authors presented an intuitive, self-explanatory interface for diabetes prediction, utilizing four ML algorithms: Decision Tree (DT), K-nearest Neighbor (KNN), Support Vector Classification (SVC), and Extreme Gradient Boosting (XGB). The authors applied these algorithms to open-source clinical data, focusing on features such as pregnancies, glucose, blood pressure, and insulin levels.

In \cite{Modak:2024}, the authors introduced a diabetes prediction model that utilized multiple ML techniques. Specifically, the study employed algorithms such as Logistic Regression, SVM, Naïve Bayes, and Random Forest, alongside various ensemble learning methods, including XGBoost, LightGBM, CatBoost, AdaBoost, and Bagging. These ensemble methods integrate predictions from multiple base learners to enhance the model's accuracy and robustness. The models were evaluated on a Kaggle dataset containing 5,000 patient records.

\textbf{Summary.} In the studies mentioned above, none of (\cite{Tasin:2023}, \cite{Orlando:2023}, \cite{Sofany:2024}, \cite{Lugner:2024}, \cite{Dharmarathne:2024}, and \cite{Modak:2024}) incorporated time-series features into their models. Furthermore, these studies focused on predicting diabetes at the individual patient level rather than across broader populations and thus, may not be suited to policy level decisions which require greater numbers to better inform decision making. Similar to our approach, the research in \cite{Lim:2023} proposed a novel progressive self-transfer framework for time series disease prediction. They used data from the Korean National Institute of Health, which included biannual medical checkup and survey information from participants aged 40 to 69 years between 2001 and 2018. The authors employed the least absolute shrinkage and selection operator (LASSO) for feature selection. However, they did not integrate datasets to obtain more comprehensive information. Notably, their work also focused on individual-level data rather than addressing predictions at the greater population level. Additionally, they introduced new features to existing models rather than improving the model itself.

\section{Our Datasets}
\label{sec:Dataset}
In addition to primary factors that increase an individual's risk of diabetes, such as glucose level, insulin level, and body mass index, we identified other potential state-level influences. These include: (1) chronic diseases such as alcohol consumption, smoking, asthma, and high cholesterol; (2) demographic factors like race and gender; (3) housing conditions, including home ownership or rental status; and (4) economic factors such as employment status, income, and poverty levels. To explore these relationships, we utilized four freely available datasets covering chronic diseases, population demographics, housing, and economic conditions across U.S. states from 2011 to 2021.

\subsection{Separate Datasets}
\subsubsection{U.S. CDI:} The chronic disease indicator (CDI) dataset was sourced from the U.S. Centers for Disease Control and Prevention (CDC) \footnote{\url{https://www.cdc.gov/}}. In collaboration with the Council of State and Territorial Epidemiologists and the National Association of Chronic Disease Directors, the CDC has developed a comprehensive suite of 115 chronic disease indicators. These indicators enable consistent definitions, data collection, and reporting across states and territories, supporting public health initiatives and allowing for uniform tracking and analysis of chronic diseases across regions.

The dataset, US-CDI Version 13 \cite{CDC-DPH}, contains 34 columns, with 24 populated by health-related data from multiple sources covering the years 2001 to 2021. It provides information on 17 distinct chronic diseases for each U.S. state and includes demographic details such as race and gender, organized by topic and specific health-related questions.

\subsubsection{U.S. population:} The second dataset, obtained from the U.S. Census Bureau\footnote{\url{https://www.census.gov}} and the U.S. National Cancer Institute\footnote{\url{https://cancer.gov/}}, provides detailed demographic information on the U.S. population. It includes race data classified into five categories: Non-Hispanic White, Non-Hispanic Black, Non-Hispanic American Indian/Alaska Native, Non-Hispanic Asian or Pacific Islander, and Hispanic. Additionally, the dataset includes data on sex (male and female) and age, organized into 19 distinct groups, covering all U.S. states.

For this paper, data covering the period from 2011 to 2021 was used. Specifically, data from 2011 to 2020 was directly extracted from the dataset provided by the U.S. National Cancer Institute \cite{NCI-24}, while data for 2021 was estimated through linear interpolation based on population trends from the preceding decade and the 2021 U.S. Census \cite{U.S.Census-24}. This approach ensures a consistent and comprehensive demographic perspective aligned with the study period.

\subsubsection{U.S. Housing:} The third data source includes housing data for each state from the years 2010 and 2020, provided by the U.S. Census Bureau via the DEC Redistricting Data (PL 94-171) product \cite{Census-24b}. This dataset contains information on the total number of housing units, as well as occupied and vacant units across all 51 states. Linear interpolation was applied to estimate housing data for the years 2011 to 2019 and 2021 based on this dataset alongside the U.S. Census dataset.

\subsubsection{U.S. Economy:} The fourth data source includes economic data for each U.S. state from 2011 to 2021, combining information from the U.S. Bureau of Labor Statistics \cite{BLS-2024}, the U.S. Census Bureau \cite{U.S.Census-24-c}, and the Bureau of Economic Analysis \cite{BEA-24}. This dataset provides details on the employment rate, per capita income, and poverty rate across states over an 11-year period.

\subsection{Our Integrated Diabetes Dataset}

\vspace{-8mm}
\begin{table}[H]
\scriptsize
\centering
\caption{Categories and their feature counts in our dataset, presented as percentages for each U.S. state by year}

\begin{tabular}{|p{1.6cm}|p{9cm}|c|}
\hline
\textbf{Category} & \textbf{Description} & \textbf{\#Fea.} \\ \hline

Diabetes & The percentage of diagnosed diabetes among adults aged 18 and older & 1 \\ \hline
Age groups & The percentage of the population, categorized by age groups. The age groups are 0-19, 20-39, 40-59 and 60+ & 4 \\ \hline
Races & The percentage of the population categorized by races. The racial groups include  Hispanic, Non-Hispanic White, Non-Hispanic Asian or Pacific Islander, Non-Hispanic American Indian or Alaska Native, and Non-Hispanic Black & 5  \\ \hline
Gender & The percentages of males and females, as well as the percentage of the overall population & 3 \\ \hline
House & The percentage of total houses, as well as The percentages of vacant and occupied houses & 3 \\ \hline

Economy & The percentage of the employed population, per capita income, and the poverty rate & 3 \\ \hline
Chronic Disease Indicators & The percentage of adults diagnosed with at least one chronic disease (excluding diabetes) or a chronic disease indicator, such as asthma, arthritis, kidney disease, high cholesterol, smoking, or having had a foot examination, dilated eye examination, or glycosylated hemoglobin measurement       & 71 \\ \hline
& \hspace{6cm} \textbf{Total Feature Count:} & \textbf{90} \\ \hline
\end{tabular}
\label{tab:dataset-category}
\end{table}

We have combined the four datasets into a unified dataset to predict the prevalence of diagnosed diabetes among adults aged 18 and older across all 51 U.S. states for the following year. This work is not covered here as the methodology is presented in earlier work in \cite{Scriney:2019} where integration from semi-structured sources was described and in \cite{10386435} where the deployment of the HL7 common model was used to integrate Covid-19 datasets.

\vspace{-0.3cm}
\begin{figure}[H]
\centering
\begin{center}
   \includegraphics[width = 12cm, height = 6cm]{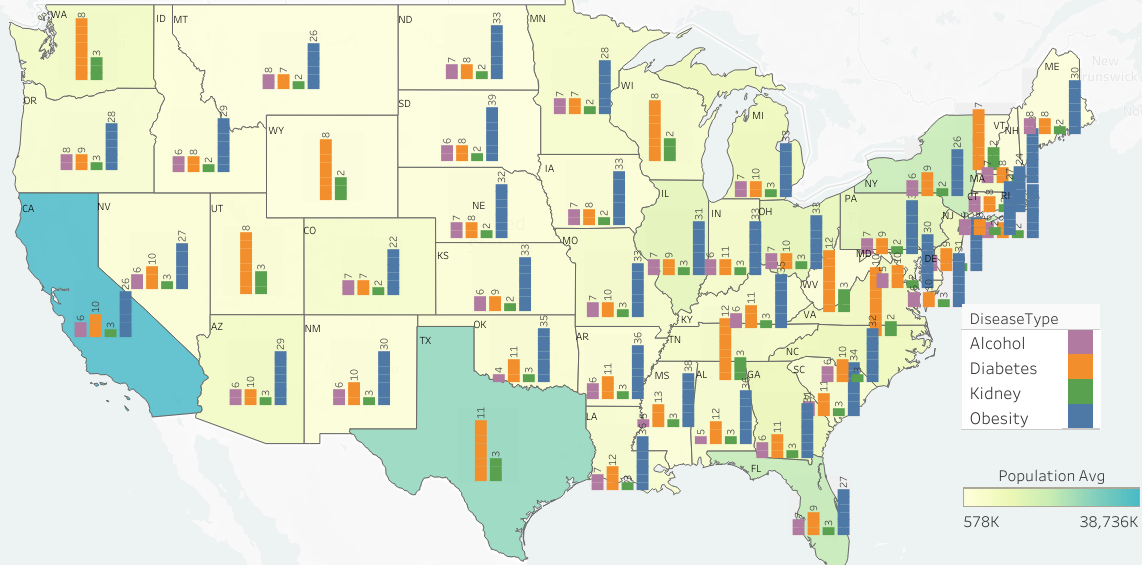}
\end{center}
\vspace{-3mm}
\caption{Average Percentage for Diabetes, Alcohol Consumption, Kidney Disease, and Obesity across U.S. states}
\label{fig_usmap_4d_group_1}
\end{figure}
\vspace{-8mm}

Table \ref{tab:dataset-category} presents seven categories together with their descriptions and feature counts in this newly formed dataset. Each feature is standardized to display the percentage of the relevant feature for each state by year. The dataset contains 90 features, including one representing the percentage of adults diagnosed with diabetes. Figure \ref{fig_usmap_4d_group_1} illustrates the average percentages over eleven years of populations in all states affected by diabetes, heavy alcohol consumption, kidney disease, and obesity.

\section{Methodology}
\label{sec:Methods}
\subsection{System Architecture}

\begin{figure}[H]
    \centering
    \captionsetup{justification=centering}
	\begin{center}
       \includegraphics[scale=0.27]{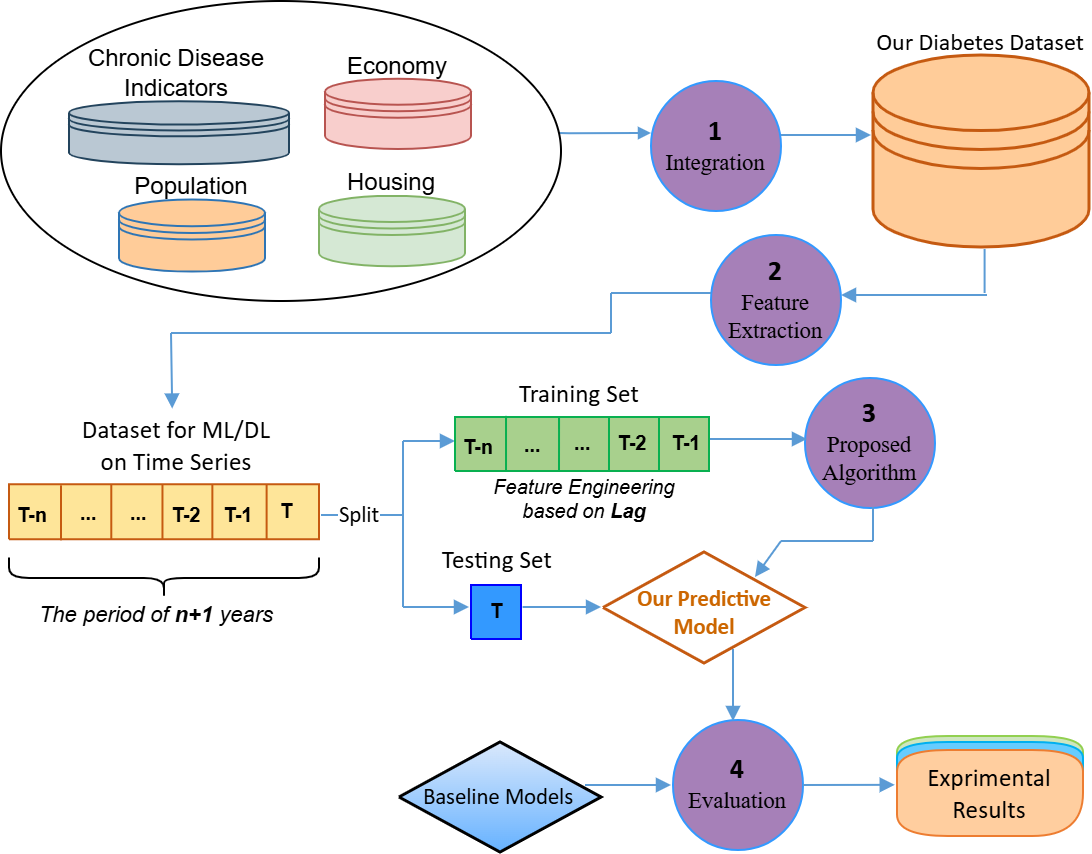}
	\end{center}
    \vspace{-2mm}
  	\caption{System Architecture: Data Engineering \& Predictive Algorithms}
	\label{fig_system_arch}
    \vspace{-5mm}
\end{figure}

The system architecture, illustrated in Figure \ref{fig_system_arch}, outlines a structured workflow for implementing our algorithm to predict diabetes trends across different states using time-series data. This architecture ensures a systematic approach to data integration, feature engineering, model development, and performance evaluation, facilitating the creation of an accurate and efficient predictive model.

The system consists of four primary processing modules. The first module, {\tt Integration}, combines various datasets into our diabetes dataset, as discussed in Section \ref{sec:Dataset}. The second module, {\tt Feature Extraction}, explains how features are used to create time-series training and testing sets, detailed in Section \ref{sec:training-testing-sets}. The third module, {\tt Proposed Algorithm}, uses the training set with engineered features based on $Lag$ to build our diabetes prediction model, described in Section \ref{sec:our_ermbag+}. Finally, the {\tt Evaluation} module (discussed in Section \ref{sec:exp_analysis}) assesses our proposed model’s predictive performance and computational efficiency on test dataset, comparing our approach with baseline models (outlined in Section \ref{sec:baseline_models}).

\subsection{Structuring Time Series Training and Testing Datasets}
\label{sec:training-testing-sets}
The objective of this study is to predict the diabetes class for a given year \(t\). The input data includes 89 predictor features for each year: 4 features related to age, 5 related to race, 3 related to gender, 3 related to housing, 3 related to economic factors, and 71 related to chronic disease indicators, as shown in Table \ref{tab:dataset-category}.

To capture the temporal dynamics of the data, the model incorporates lagged observations from previous years. For a given $Lag$ \( l \), where $l = 1$ to $9$, the input dataset consists of \( 89 \times (l + 1) + l\) variables. This includes 89 predictor variables for the current year \( t \) and additional sets of 89 predictors for each of the preceding years up to \( t-l \), along with \( l \) corresponding target output values for years \( t-1 \) through \( t-l \).

For example, with $Lag = 1$, there are 179 variables in total: 89 input variables for year \( t \), 89 for year \( t-1 \), and 1 output variable from year \( t-1 \). For $Lag = 2$, the model includes 269 variables, comprising 89 input variables each for years \( t \), \( t-1 \), and \( t-2 \), plus 2 output values for years \( t-1 \) and \( t-2 \). With $Lag = 9$, the model aggregates 899 variables, including 9 past output values and 890 input variables spanning years \( t-9 \) through \( t \).

The training set includes observations from 2011 to 2020, while the test set comprises data for 2021. The test set consistently contains 51 observations, corresponding to the 51 states to be forecasted for 2021. In contrast, the number of observations in the training set varies depending on the lag years considered.  For each $Lag = l$, the number of training observations is calculated as $51 \times (10 - l)$, covering the period from $2011 + l$ through 2020. Consequently, the number of training observations for each $Lag$ from 1 to 9 is 459, 408, 357, 306, 255, 204, 153, 102, and 51, respectively.

\subsection{Popular Time Series Prediction Models}
\label{sec:baseline_models}
The six machine learning models listed below are widely used in healthcare applications for their ability to capture complex patterns in medical data, including disease progression, patient monitoring, and predictive analytics.

\subsubsection{a) Support Vector Machine Regression (SVMReg):}
SVMReg extends Support Vector Machines for regression by finding a function that minimizes prediction errors within a margin \(\epsilon\) \cite{Awad-Other:2015}. The model is defined as:
\begin{equation}
    f(x) = \langle w, \phi(x) \rangle + b
\end{equation}
where \(w\) is the weight vector, \(\phi(x)\) maps input \(x\) into a higher-dimensional space, and \(b\) is the bias.

\subsubsection{b) Binary Decision Tree (BDTree):}
BDTree recursively partitions the input space to predict continuous values \cite{Silva-Others:2021}. It minimizes variance by selecting feature \(j\) and threshold \(t_j\) that optimize:
\begin{equation}
\min_{j, t_j} \sum_{i \in \text{left}} (y_i - \bar{y}_\text{left})^2 + \sum_{i \in \text{right}} (y_i - \bar{y}_\text{right})^2
\end{equation}
where \(\bar{y}_\text{left}\) and \(\bar{y}_\text{right}\) are the mean target values of the child nodes.

\subsubsection{c) Least-Squares Boosting (LSBoost):}
LSBoost enhances regression accuracy by iteratively fitting decision trees to residuals \cite{Friedman-Others:2001}. The final prediction is:
\begin{equation}
\hat{y} = \sum_{m=1}^{M} \beta_m h_m(x)
\end{equation}
where \(h_m(x)\) represents the \(m\)-th tree and \(\beta_m\) are weights.

\subsubsection{d) Neural Network (NN)}
NNs model complex patterns using interconnected neurons across layers \cite{he2021neural}. A multi-layer NN follows:
\begin{equation}
\hat{y} = \phi_L(\phi_{L-1}(\cdots \phi_1(W_1 x + b_1) \cdots ) + b_L)
\end{equation}
where \(W_l\) and \(b_l\) are weights and biases, and \(\phi_l(\cdot)\) is the activation function.

\subsubsection{e) Long Short-Term Memory (LSTM):}
LSTM addresses long-term dependencies in sequential data using memory cells \cite{men2021multi}. It updates states via:
\begin{equation}
\begin{aligned}
    f_t &= \sigma(W_f x_t + U_f h_{t-1} + b_f) \\
    i_t &= \sigma(W_i x_t + U_i h_{t-1} + b_i) \\
    \tilde{C}_t &= \tanh(W_c x_t + U_c h_{t-1} + b_c) \\
    C_t &= f_t \odot C_{t-1} + i_t \odot \tilde{C}_t \\
    o_t &= \sigma(W_o x_t + U_o h_{t-1} + b_o) \\
    h_t &= o_t \odot \tanh(C_t)
\end{aligned}
\end{equation}
where \(\sigma\) and \(\tanh\) are activation functions, \(\odot\) denotes element-wise multiplication, and \(W, U, b\) are model parameters.

\subsubsection{f) Bagging Ensemble Regression (ERMBag):}
ERMBag is a powerful method for improving the stability and accuracy of regression predictions by combining multiple models \cite{Zhao-Others:2023}. The technique begins by creating multiple bootstrap samples from the original dataset \(\{(x_i, y_i)\}_{i=1}^{n}\), where each sample is generated through random sampling with replacement. Each of these bootstrap samples is used to train an individual regression tree model, resulting in a collection of \(B\) different models.

The predictions from these individual models are then aggregated to produce a final prediction. The most common method of aggregation is by averaging the predictions:

\begin{equation}
\hat{y} = \frac{1}{B} \sum_{b=1}^{B} f_b(x)
\end{equation}

where \(\hat{y}\) is the final prediction for a given input \(x\), and \(f_b(x)\) represents the prediction of the \(b\)-th model.

\subsection{Our ERMBag+}
\label{sec:our_ermbag+}
Among the six ML models mentioned above, ERMBag is particularly popular for its robustness and ensemble-based approach, which improves predictive accuracy by reducing variance. In the context of disease prediction in time series, ERMBag offers several advantages. By aggregating multiple regression models, it enhances stability and resilience against overfitting, making it well-suited for handling noisy and imbalanced medical datasets. Additionally, ERMBag can effectively capture temporal dependencies in disease progression, improving early detection and long-term forecasting accuracy. Its adaptability to different feature sets and an ability to integrate various weak learners further enhance its predictive power in healthcare applications.

However, ERMBag still has some disadvantages that need improvement. To address these issues, we propose ERMBag+, as described in Algorithm \ref{alg:ermgbag+}, to enhance the model's robustness and predictive accuracy. This is achieved by refining resampling methods, optimizing prediction aggregation, and implementing adaptive strategies tailored to the dataset's characteristics. 

\begin{algorithm}[H]
\caption{Our ERMBag+: Ensemble Bagging-Based Regression using Decision tree for Diabetes Prevalence Forecasting}
\label{alg:ermgbag+}
\begin{algorithmic}[1]
\Require 
    \Statex Time series dataset $D = \{(X_t, y_t)\}$, where $X_t$ represents the feature vector and $y_t$ denotes the target variable (diabetes prevalence).  
    \Statex Number of base learners: $M$  
    \Statex Block size for bootstrapping: $B$  
    \Statex Base model: Decision Trees  
    \Statex Lag parameter: $l \in \{1,2,\dots,9\}$  
    \Statex Performance evaluation metric: Root Mean Square Error (RMSE)  
\Ensure 
    \Statex Final ensemble prediction $\hat{y}$ for all 51 states in the year 2021.  

\State \textbf{Step 1: Data Preprocessing}  
    \State Transform the dataset into a supervised learning format:
    \begin{align*}
        X_t &= [X_t, X_{t-1}, \dots, X_{t-l}, y_{t-1}, \dots, y_{t-l}]
    \end{align*}
    \State Define the total number of predictive features:  
        $d = 89 \times (l+1) + l$
    \State Normalize all input variables.  

\State \textbf{Step 2: Stratified Block Bootstrap Sampling}  
    \State Partition the dataset into non-overlapping time blocks of size $B$.  
    \State Ensure stratification by preserving the distribution across demographic and economic variables.  
    \State Generate $M$ bootstrapped datasets $D_1, D_2, \dots, D_M$.  

\State \textbf{Step 3: Training an Ensemble of Decision Trees}  
\For{$i = 1$ to $M$}
    \State Train a \textbf{Decision Tree} model $f_i$ using bootstrapped dataset $D_i$.  
    \State Apply an early stopping criterion based on RMSE to mitigate overfitting.  
    \State Store the trained model $f_i$.  
\EndFor  

\State \textbf{Step 4: Adaptive Model Aggregation via Weighted Averaging}  
\For{$i = 1$ to $M$}
    \State Compute individual model predictions: $\hat{y}_{i} = f_i(X_{\text{test}})$.  
    \State Compute weight for model $f_i$: $ w_i = \frac{1}{\text{RMSE}(f_i)} $
\EndFor  
\State Normalize model weights: $w_i^* = \frac{w_i}{\sum_{j=1}^{M} w_j}$
\State Compute final ensemble prediction: $ \hat{y} = \sum_{i=1}^{M} w_i^* \hat{y}_{i} $

\end{algorithmic}
\end{algorithm}

Specifically, given the presence of lagged features in the dataset, it is crucial to adjust the block size ($B$) in the bootstrap process ({\tt Step 1} of the algorithm) to align with the data’s time-based dependencies. For datasets with significant lag, increasing the block size helps preserve temporal patterns, ensuring more accurate predictions. Conversely, for datasets with minimal lag, reducing the block size introduces greater variability, improving the ensemble’s ability to generalize across different scenarios.

Instead of relying on a simple bootstrap approach, we recommend adopting the Stratified Block Bootstrap technique ({\tt Step 2} of the algorithm). This method ensures that resampled datasets accurately reflect the distribution of specific subgroups—such as those defined by state or time period—thereby reducing potential biases caused by uneven subgroup representations.

To mitigate overfitting, particularly in high-dimensional datasets prone to multicollinearity, we integrate an early stopping criterion for each model in the ensemble ({\tt Step 3} of the algorithm). This mechanism monitors validation error rates and halts training once improvements plateau, thereby preserving the model's ability to generalize to new data.

To further improve the accuracy of combined predictions, we propose a weighted aggregation framework, as outlined in {\tt Step 4} of the algorithm. In this approach, each decision tree's contribution is adjusted based on its performance on validation datasets. Specifically, weights are assigned inversely proportional to each model's Root Mean Square Error (RMSE), ensuring that models with lower errors have a greater impact on the final prediction.

These enhancements address key challenges like data imbalance, overfitting, and temporal dependencies in time-series and high-dimensional datasets. By integrating adaptive resampling, weighted aggregation, and early stopping, the model improves stability, accuracy, and generalization. Additionally, these improvements enhance the model’s ability to generalize across different datasets, reducing the risk of bias and improving robustness in real-world healthcare applications. Ultimately, these refinements contribute to a more reliable and interpretable predictive framework, making it well-suited for complex decision-making tasks in dynamic environments.

\section{Experiments and Analysis}
\label{sec:exp_analysis}

This section corresponds to Modules 3 and 4 in Figure \ref{fig_system_arch}. For ERMBag+ and all six baseline machine learning models, we present and analyze both their predictive performance and computational efficiency.

\subsection{Measures}

To assess the performance of prediction models, several widely-used metrics include the Mean Absolute Error, $MAE = \frac{1}{n} \sum_{i=1}^{n} |y_i - \hat{y}_i|$  \cite{Willmott-Other:2005}; Root Mean Squared Error, $\text{RMSE} = \sqrt{\frac{1}{n} \sum_{i=1}^{n} (y_i - \hat{y}_i)^2}$ \cite{Hodson-Other:2022}; the Mean Absolute Percentage Error, $\text{MAPE} = \frac{100\%}{n} \sum_{i=1}^{n} \left| \frac{y_i - \hat{y}_i}{y_i} \right|$ \cite{deMyttenaere-Other:2016}; and the coefficient of determination, $R^2 = 1 - \frac{\sum_{i=1}^{n} (y_i - \hat{y}_i)^2}{\sum_{i=1}^{n} (y_i - \bar{y})^2}$ \cite{Chicco-Other:2021}. where \(y_i\) is the actual value, \(\hat{y}_i\) is the predicted value, and \(n\) is the number of observations. Each of these metrics offers a distinct perspective on the model's prediction accuracy.

The algorithms were executed using MATLAB 2022b\footnote{\url{https://se.mathworks.com/products/new_products/release2022b.html}} on a 64-bit Windows 11 system. All experiments were conducted on a computer equipped with an AMD Ryzen 7 5700U CPU with Radeon Graphics and 24GB of RAM.

\subsection{Results and Discussion}

\begin{figure}[H]
    \begin{center}
    \includegraphics[width = 0.9\textwidth]{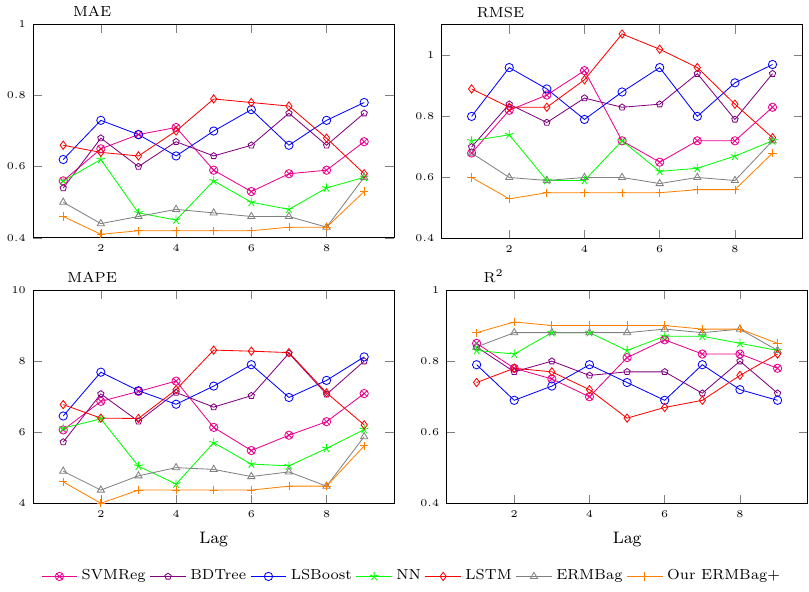}
    \end{center}
    \vspace{-3mm}
    \caption{Average prediction performance of models}
    \vspace{-3mm}
    \label{fig:Exp_Result}
\end{figure}

Figure \ref{fig:Exp_Result} presents the average prediction performance of six models—SVMReg, BDTree, LSBoost, ERMBag, NN, and LSTM—evaluated across $Lag$ values from 1 to 9, using the metrics MAE, RMSE, R$^2$, and MAPE. The NN model is configured with two layers, each containing 10 hidden neurons, while the LSTM model employs a dropout rate of 0.005 and a maximum of 500 training epochs. The results highlight each model's optimal performance across different $Lag$ values, as summarized in Table \ref{table:model_performance}. Specifically, BDTree and LSBoost achieve their best performance at $Lag = 1$, while ERMBag and our enhanced ERMBag+ perform best at $Lag = 2$. Meanwhile, NN, SVMReg and LSTM reach their optimal performance at $Lag$ values of 4, 6, and 9, respectively.

\vspace{-6mm}

The experimental results in Table \ref{table:model_performance} demonstrate varying performance across models in terms of predictive accuracy and computational efficiency. Our \-ERMBag+ achieves the best overall performance at $Lag = 2$, with the lowest error rates (MAE = 0.41, RMSE = 0.53, MAPE = 4.01) and the highest predictive accuracy (R$^2$ = 0.91). However, this requires a substantial computational cost for training.

Among the baseline models, ERMBag ($Lag = 2$) is the second-best performer, achieving MAE = 0.44, RMSE = 0.60, MAPE = 4.38, and R$^2$ = 0.88, while maintaining a more reasonable total execution time of 1.591 seconds. The NN model ($Lag = 4$) also shows strong predictive ability with R$^2$ = 0.88, MAE = 0.45, RMSE = 0.59 and MAPE = 4.55, but its training time (8.182 seconds) is significantly higher than ERMBag.

Other models demonstrate moderate performance with trade-offs between accuracy and efficiency. SVMReg ($Lag = 6$) achieves R$^2$ = 0.86, MAE = 0.53, and RMSE = 0.65, while maintaining a fast total execution time of 0.076 seconds. BDTree and LSBoost (both optimized at $Lag = 1$) perform well for short-term predictions but exhibit relatively higher errors (MAE = 0.54 and 0.62, respectively). LSTM ($Lag = 9$) has an R$^2$ of 0.82 but requires 11.345 seconds for execution, making it the slowest models.

\vspace{-5mm}
\begin{table}[H]
\centering
\caption{Model Performance Summary}
\begin{tabular}{|l|c|c|c|c|c|c|c|c|}
\hline
\multirow{2}{*}{\textbf{Model}} & \multirow{2}{*}{\textbf{Lag}} & \multicolumn{4}{c|}{\textbf{Prediction Performance}} &  \multicolumn{3}{c|}{\textbf{Time Performance (second)}} \\ \cline{3-9}
 & & \textbf{MAE} & \textbf{RMSE} & \textbf{MAPE} & \textbf{R\textsuperscript{2}}  & Training & Prediction & Total
 \\ \hline
 \rowcolor{LightYellow}
 LSBoost & 1 & 0.62 & 0.80 & 6.46 & 0.79 & 0.598 & 0.032 & 0.63 \\ \hline
 \rowcolor{LightYellow}
 LSTM & 9 & 0.58 & 0.73 & 6.21 & 0.82 & 11.309 & 0.036 & 11.345 \\ \hline

 \rowcolor{LightCyan}
 BDTree & 1 & 0.54 & 0.70 & 5.73 & 0.84 & 0.045 & 0.011 & 0.056 \\ \hline
 \rowcolor{LightCyan}
 SVMReg & 6 & 0.53 & 0.65 & 5.49 & 0.86 & 0.049 & 0.027 & 0.076 \\ \hline

 \rowcolor{lime}
 NN & 4 & 0.45 & 0.59 & 4.55 & 0.88 & 8.182 & 0.013 & 8.195 \\ \hline
 \rowcolor{lime}
 ERMBag & 2 & 0.44 & 0.6 & 4.38 & 0.88 & 1.537 & 0.054 & 1.591 \\ \hline

 \rowcolor{orange}
 ERMBag+ & 2 & 0.41 & 0.53 & 4.01 & 0.91 & 3.235 & 0.029 & 3.264 \\ \hline
\end{tabular}
\label{table:model_performance}
\end{table}
\vspace{-5mm}

Overall, ERMBag+ achieves the highest predictive performance, surpassing the second-best models (ERMBag and NN) by approximately 3.4\% to 13.5\% across various evaluation metrics. Additionally, it outperforms the third-best models (SVMReg and BDTree) by 5.8\% to 42.9\% and exceeds the weakest performing models (LSTM and LSBoost) by 11\% to 61.1\%.

\section{Conclusion and Future Work}
\label{sec:Conclusion}

We integrated multiple U.S. diabetes-related datasets to construct a comprehensive dataset containing valuable information on diabetes, collected for each state and year from 2011 to 2021. This dataset enables accurate state-level predictions and provides deeper insights into regional trends in diabetes prevalence. Specifically, we proposed EBMBag+, an enhancement of the traditional EBMBag model that incorporates time series techniques for improved diabetes prediction. Our experimental results demonstrated that EBMBag+ achieved the highest predictive performance, outperforming baseline models—LSBoost, LSTM, BDTree, SVMReg, NN, and ERMBag—by 3.4\% to 61.1\% across the R$^2$, MAE, RMSE, and MAPE metrics. This approach not only improves prediction accuracy but also offers valuable insights into the evolving dynamics of diabetes at the state level, empowering policymakers to make more informed decisions and implement targeted healthcare strategies.

Future research will see additional features related to diabetes in states  including healthcare infrastructure, environmental pollutants, climate and weather, lifestyle factors, and socioeconomic status. In addition, we will apply feature selection techniques to evaluate which features in our dataset have the most significant impact on the performance of our models.

\section*{Acknowledgement}
The research is part of the RECONNECT project: chRonic disEase: disCOvery, aNalysis aNd prEdiCTive modelling. This publication has emanated from research conducted with the financial support of Taighde Éireann – Research Ireland under Grant numbers 22/NCF/DR/11244 and 12/RC/2289\_P2.

\bibliographystyle{splncs04}
\bibliography{References.bib}

\end{document}